\documentclass[10pt, a4paper]{article}
\usepackage{algorithm}
\usepackage{algorithmicx}
\usepackage{algpseudocode}
\usepackage{amsmath}
\usepackage{multicol}
\usepackage{url}
\usepackage{tabularray}
\usepackage{booktabs}
\usepackage{multirow}
\usepackage[final]{lrec2026} 

\title{AstroConcepts: A Large-Scale Multi-Label Classification Corpus for Astrophysics}

\name{Atilla Kaan Alkan$^1$, Felix Grezes$^1$, Sergi Blanco-Cuaresma$^{1,}$$^2$, \\ \large \textbf{Jennifer Lynn Bartlett$^1$}, \textbf{Daniel Chivvis$^1$}, \textbf{Anna Kelbert$^1$}, \\ \large \textbf{Kelly Lockhart$^1$}, \textbf{Alberto Accomazzi}$^1$\\[0.05em]}

\address{$^1$Harvard-Smithsonian Center for Astrophysics, Cambridge, MA, USA \\
         $^2$Faculty of Psychology, UniDistance Suisse, Brig, Switzerland \\[0.05em]
\{atilla.alkan, felix.grezes, sblancocuaresma, jennifer.bartlett,\\ daniel.chivvis, anna.kelbert, kelly.lockhart, alberto.accomazzi\}@cfa.harvard.edu\\}

\abstract{
Scientific multi-label text classification suffers from extreme class imbalance, where specialized terminology exhibits severe power-law distributions that challenge standard classification approaches. Existing scientific corpora lack comprehensive controlled vocabularies, focusing instead on broad categories and limiting systematic study of extreme imbalance. We introduce \textsc{AstroConcepts}, a corpus of English abstracts from 21,702 published astrophysics papers, labeled with 2,367 concepts from the Unified Astronomy Thesaurus. The corpus exhibits severe label imbalance, with 76\% of concepts having fewer than 50 training examples. By releasing this resource, we enable systematic study of extreme class imbalance in scientific domains and establish strong baselines across traditional, neural, and vocabulary-constrained LLM methods. Our evaluation reveals three key patterns that provide new insights into scientific text classification. First, vocabulary-constrained LLMs achieve competitive performance relative to domain-adapted models in astrophysics classification, suggesting a potential for parameter-efficient approaches. Second, domain adaptation yields relatively larger improvements for rare, specialized terminology, although absolute performance remains limited across all methods. Third, we propose frequency-stratified evaluation to reveal performance patterns that are hidden by aggregate scores, thereby making robustness assessment central to scientific multi-label evaluation. These results offer actionable insights for scientific NLP and establish benchmarks for research on extreme imbalance.
\\ \newline 
}

\begin{document}

\maketitleabstract

\section{Introduction}

Scientific multi-label text classification poses significant challenges due to extreme class imbalance, with specialized terminology exhibiting severe power-law distributions that challenge standard classification methods~\citep{liu-etal-survey}. While imbalanced distributions are common in scientific domains, existing corpora~\citep{giles-etal-1998,McCallum2000AutomatingTC,yang-etal-2018-sgm} typically provide limited coverage of controlled vocabulary, focus on broad disciplinary categories, or operate at scales that hinder systematic methodological investigation of extreme imbalance scenarios. This limitation is particularly problematic for comprehensive scientific text classification, where controlled vocabularies organize thousands of specialized concepts with naturally occurring power-law distributions. Existing datasets that provide controlled-vocabulary coverage either operate at prohibitive computational scales~\citep{toney-dunham-2022-multi} or focus on limited subsets of domain terminology, thereby preventing systematic investigation of how different approaches address the fundamental challenge of learning from severely imbalanced specialized terminologies.

The astrophysics literature exemplifies these challenges while providing an ideal testbed for systematic investigation. The Unified Astronomy Thesaurus (UAT)~\citep{accomazzi-etal-uat} organizes 2,367 concepts across 11 hierarchical levels, creating a comprehensive controlled vocabulary that exhibits natural power-law distributions characteristic of scientific domains.

To provide the community with essential resources for investigating extreme multi-label classification in scientific domains, we introduce \textsc{AstroConcepts}, a corpus of 21,702 astrophysics papers labeled with the complete UAT vocabulary (2,367 concepts). This resource enables systematic investigation of three fundamental research questions:

\begin{itemize}
\item \textbf{RQ1} How do traditional methods, supervised neural models, and vocabulary-constrained LLMs compare for handling extreme label imbalance in scientific classification?
\item \textbf{RQ2} Does domain-specific pretraining provide uniform benefits across frequency bins, or do improvements concentrate in specific regions of the label distribution?
\item \textbf{RQ3} Can vocabulary-constrained LLMs achieve competitive performance with domain-adapted models in astrophysics classification?
\end{itemize}

Our main contributions are:

\begin{enumerate}
    \item \textsc{AstroConcepts}\footnote{\url{https://huggingface.co/datasets/adsabs/SciX_UAT_keywords}} provides the community with the first tractable-scale corpus, enabling systematic investigation of extreme multi-label classification with comprehensive controlled-vocabulary coverage.
    
    \item Systematic comparison across traditional, neural, and vocabulary-constrained LLM approaches reveals competitive performance for parameter-efficient methods, opening new research directions for scientific NLP.
    
    \item Introduction of frequency-stratified evaluation framework with robustness metrics that reveal performance patterns invisible in aggregate scores, providing essential tools for extreme multi-label assessment.
    
    \item Demonstration that domain adaptation improvements concentrate on rare specialized terminology, with implications for training strategies in scientific applications.
    
    \item Comprehensive baseline establishment that provides essential benchmarks for scientific multi-label classification while revealing fundamental properties of extreme imbalance in specialized domains.
\end{enumerate}


The remainder of this paper is structured as follows. Section~\ref{related_work} reviews related work in scientific multi-label classification and positions \textsc{AstroConcepts} within the existing landscape of scientific corpora. Section~\ref{corpus_description} describes the corpus construction methodology, annotation procedures, and key characteristics of the resulting dataset. Section~\ref{experiments} details our experimental setup, including baseline methods, evaluation metrics, and a frequency-stratified analysis framework. Section~\ref{results_analysis} presents results across all methods, revealing key patterns in overall performance and frequency-specific behaviors. Section~\ref{discussion} discusses the broader implications of our findings for scientific NLP, methodological contributions, and limitations of the current work. Section~\ref{conclusion} concludes with a summary of key insights and directions for future research.

\section{Scientific Multi-label Classification Benchmarks}
\label{related_work}


Multi-label text classification spans diverse domains and scales. Early work established evaluation protocols with news categorization~\citep{reuters-21578_text_categorization_collection_137}, while legal document classification introduced a hierarchical structure through EUR-Lex~\citep{LozaMencia2010}, though annotations include complete paths rather than the flat terminal concepts typical in practice. Extreme multi-label benchmarks (Amazon-670K, Wiki-500K) scale to hundreds of thousands of labels but prioritize computational efficiency over domain-specific controlled vocabularies.

Scientific classification began with computer science papers: \citet{giles-etal-1998} used six broad categories, \citet{McCallum2000AutomatingTC} expanded to 70 categories with a 3-level hierarchy, and later work incorporated controlled vocabularies \citep{Santos2009MultilabelHT,Kowsari2017HDLTexHD,yang-etal-2018-sgm}, although these remained limited in scope. Recent efforts leverage Microsoft Academic Graph: \citet{cohan-etal-2020-specter} created embeddings across 19 fields, \citet{sadat-caragea-2022-hierarchical} scaled to 186K papers with 6-level hierarchy, and \citet{toney-dunham-2022-multi} used 180--220M papers, though computational demands limit systematic experimentation. While MAG provides broad coverage, it spans multiple disciplines rather than offering deep domain specialization.

Table~\ref{tab:benchmarks} puts \textsc{AstroConcepts} in context of existing resources. General benchmarks lack controlled vocabularies, whereas scientific corpora either use ad hoc categories, span multiple disciplines without deep specialization, or reach a prohibitive scale. \textsc{AstroConcepts} uniquely combines tractable scale (21K), deep hierarchy (11 levels), and domain-specific controlled vocabulary (UAT), enabling comprehensive evaluation of hierarchy-aware methods and few-shot learning in a realistic scientific classification scenario.

\begin{table*}[t]
\centering
\small
\begin{tabular}{@{}lrrrrrr@{}}
\toprule
\textbf{Corpus} & \textbf{Docs} & \textbf{Labels} & \textbf{Hierarchy} & \textbf{Supervision} & \textbf{Vocabulary} & \textbf{Domain} \\
\midrule
\multicolumn{7}{@{}l}{\emph{General Domain}} \\
\citet{LozaMencia2010} & 19K & 7.2K & 2-level & Hierarchical & EuroVoc & Legal \\
\citet{reuters-21578_text_categorization_collection_137} & 21K & 90 & None & Flat & Ad-hoc & News \\
RCV1 & 800K & 103 & 4-level & Flat & Ad-hoc & News \\
\addlinespace[0.5ex]
\multicolumn{7}{@{}l}{\emph{Scientific Domain}} \\
\citet{giles-etal-1998} & 3K & 6 & None & Flat & Ad-hoc & CS \\
\citet{Santos2009MultilabelHT} & 15K & 92 & 2-level & Mixed & ACM & CS \\
\textbf{\textsc{AstroConcepts}} & \textbf{21K} & \textbf{2.3K} & \textbf{11-level} & \textbf{Flat} & \textbf{UAT} & \textbf{Astrophys} \\
\citet{cohan-etal-2020-specter} & 25K & 19 & 1-level & Flat & MAG & Multi-sci \\
\citet{Kowsari2017HDLTexHD} & 47K & 134 & 2-level & Hierarchical & WoS & CS+Med \\
\citet{McCallum2000AutomatingTC} & 53K & 70 & 3-level & Hierarchical & Ad-hoc & CS \\
\citet{yang-etal-2018-sgm} & 55.8K & 54 & 2-level & Flat & Ad-hoc & CS \\
\citet{sadat-caragea-2022-hierarchical} & 186K & 1.2K & 6-level & Mixed & MAG & Multi-sci \\
\citet{toney-dunham-2022-multi} & 180M & 313 & 2-level & Flat & MAG & Multi-sci \\
\bottomrule
\end{tabular}
\caption{Multi-label classification benchmarks sorted by corpus size. \textsc{AstroConcepts} provides tractable scale (21K documents) with deep hierarchical structure (11 levels) and domain-specific controlled vocabulary (UAT), enabling comprehensive experimentation on astrophysics literature classification.}
\label{tab:benchmarks}
\end{table*}

\section{The AstroConcepts Corpus}
\label{corpus_description}

AstroConcepts exhibits characteristics that make scientific multi-label classification challenging: a hierarchical controlled vocabulary (UAT) with flat expert annotations, severe label imbalance, and domain-specific terminology that requires specialized language understanding. This section describes the data collection process, the UAT taxonomy structure, the annotation methodology, and the comprehensive corpus statistics.

\subsection{Source Data}
We collected 21,702 abstracts from published English-language papers indexed by the NASA-funded Science Explorer (SciX;~\citet{2025AAS...24544204B}), an expansion of the Astrophysics Data System (ADS;~\citet{accomazzi-2015-ads}) to cover all NASA science disciplines. Building on the ADS legacy, SciX is the primary bibliographic database for astronomy and astrophysics. It indexes papers from approximately 8,000 refereed journals, including the \textit{Astrophysical Journal} (\textit{ApJ}). Starting in 2018, journal editors began requiring or encouraging UAT concept assignment during submission to promote standardized concept usage in astronomy, resulting in a growing corpus of controlled-vocabulary annotations.

We selected papers meeting the following criteria: (1) published in journals where authors assign UAT concepts during submission, ensuring controlled standard concept annotation, (2) at least one UAT concept assigned, and (3) English-language abstracts. The temporal scope covers 2018-2023, reflecting the data available during corpus construction. Future work could extend coverage to more recent publications. This process yielded 21,702 documents.

\subsection{The Unified Astronomy Thesaurus}
The UAT \citep{accomazzi-etal-uat} is a community-maintained controlled vocabulary for astronomical literature, owned and openly licensed by the \textit{American Astronomical Society}. It follows the Simple Knowledge Organization System (SKOS;~\citet{skos-reference}) standard and incorporates terms from earlier astronomy thesauri. Astronomy librarians and domain experts have contributed to its development and maintenance. Released in 2017 and subsequently adopted by major journals and SciX, the UAT provides standardized terminology for indexing and retrieval of astronomy content.

The UAT version 5.1.0\footnote{\url{https://github.com/astrothesaurus/UAT/tree/v.5.0.0}} used in this work contains 2,367 astronomical concepts organized in an eleven-level hierarchical structure forming a directed acyclic graph (DAG). Concepts range from broad top-level categories such as cosmology or observational astronomy, to highly specific concepts such as the Kreutz group. Each concept includes a unique identifier, a canonical designation, alternative names (synonyms, acronyms), and an optional textual definition (scope note), and explicit relationships to parents, children, and related concepts.

The UAT follows a polyhierarchical structure in which concepts can have multiple parents, creating a DAG topology in which nodes (concepts) can be reached through multiple paths without forming cycles. For example, \textit{Stellar atmosphere} appears under both \textit{Stars} and \textit{Spectroscopy}, as studies of stellar atmospheres involve both stellar physics and spectroscopic methods. This DAG structure reflects the multidisciplinary nature of astronomical research, where concepts naturally belong to multiple semantic categories simultaneously.

\subsection{Concept Assignment Process}
Labels in \textsc{AstroConcepts} come from the standard publishing process where authors assign UAT concepts to their manuscripts during submission. Authors typically select 4 concepts per paper (see Table~\ref{tab:corpus_stats}) using the journal's submission system. They choose specific concepts without marking hierarchical paths. For example, selecting \textit{Exoplanet atmospheres} does not require selecting its broader categories, such as \textit{Exoplanet astronomy}. This creates an interesting challenge: systems must predict specific concepts from a hierarchical vocabulary using only flat annotations. The approach has clear advantages: we collect high-quality labels from domain experts who know their work best, and the standardized UAT vocabulary ensures consistency. However, authors naturally focus on what they consider most important rather than providing comprehensive coverage. This means some relevant concepts might be missing, creating a realistic yet challenging evaluation scenario that mirrors real-world conditions in which systems operate with an incomplete set of annotations.

\subsection{Overall Statistics}

Table~\ref{tab:corpus_stats} presents overall corpus statistics. \textsc{AstroConcepts} contains 21,702 abstracts with 93,547 total label assignments, averaging 4.31 labels per abstract. The assigned label space comprises 1,864 unique UAT concepts (92\% of the UAT vocabulary), indicating that the selected literature abstracts span the conceptual space defined by UAT. 

\begin{table}[!h]
\centering
\small
\begin{tabular}{@{}lr@{}}
\toprule
\textbf{Statistic} & \textbf{Value} \\
\midrule
\multicolumn{2}{@{}l}{\textit{Documents}} \\
\quad Total abstracts & 21,702 \\
\addlinespace[0.5ex]
\multicolumn{2}{@{}l}{\textit{Labels}} \\
\quad Total assignments & 93,547 \\
\quad Assigned unique labels & 1,864 \\
\quad Per abstract (mean/median) & 4.31 / 4 \\
\quad Per abstract (range) & 1--12 \\
\addlinespace[0.5ex]
\multicolumn{2}{@{}l}{\textit{Text}} \\
\quad Length in words (mean/median) & 211.2 / 223 \\
\quad Length (range) & 18--462 \\
\quad Vocabulary size & 163,326 \\
\bottomrule
\end{tabular}
\caption{Overall statistics for \textsc{AstroConcepts}.}
\label{tab:corpus_stats}
\end{table}

Abstracts average 211 words (median: 223), typical for scientific abstracts and compatible with standard transformer context windows (512 tokens). The distribution is approximately normal with slight positive skew toward longer abstracts; 94\% fit within 512 tokens. We retain abstracts up to 462 words to capture the full range of scientific writing styles without truncation artifacts.

The distribution of labels per abstract (mean: 4.31, median: 4, range: 1--12) reflects the multifaceted nature of astrophysics research. Most abstracts (70\%) receive 2--5 labels, with single-label papers typically representing narrowly focused studies and papers with 6+ labels (15\%) covering interdisciplinary or methodologically diverse work. This moderate label density exceeds general multi-label benchmarks such as Reuters-21578~\citep{reuters-21578_text_categorization_collection_137}, which averages 1.2 labels per document~\citep{huang-etal-2021-balancing}, underscoring the conceptual complexity inherent in scientific literature.

\subsection{Concept Frequency Distribution}

\begin{figure}[!h]
\centering
\includegraphics[width=0.99\columnwidth]{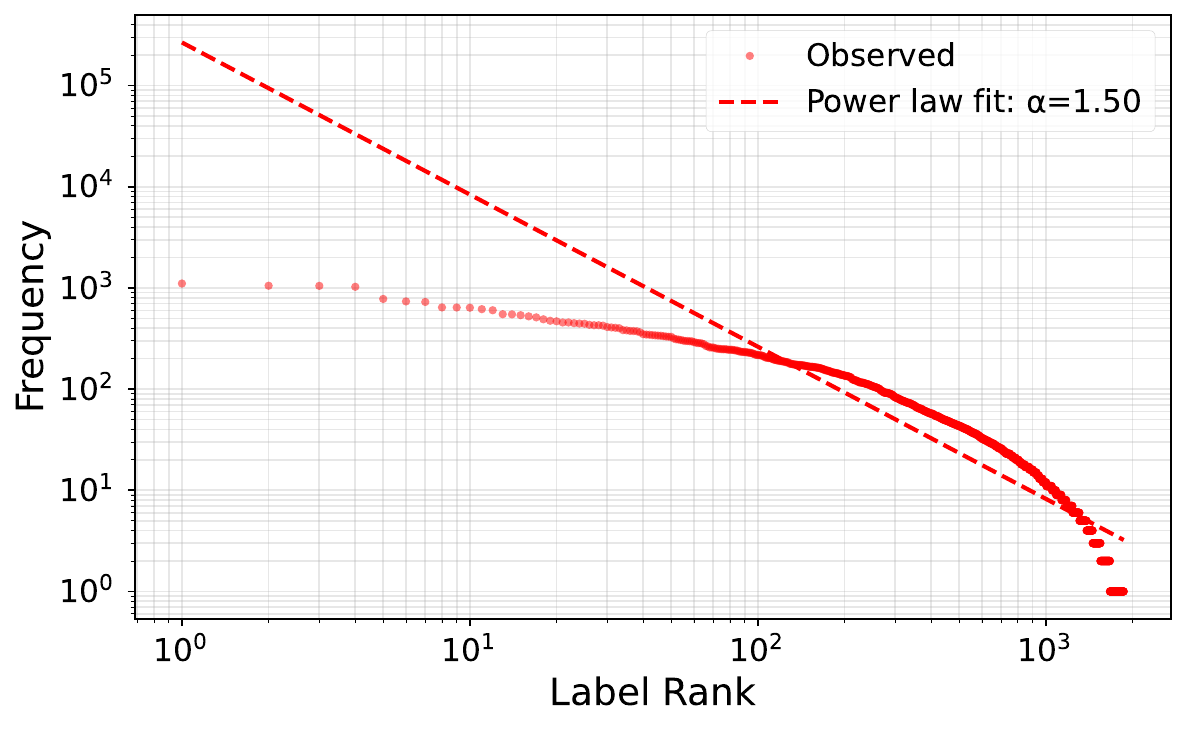}
\caption{Label frequency distribution (log-log scale) and fitted power-law function with exponent $\alpha = 1.50$ ($R^2 = 0.825$). The long tail contains 76\% of labels with fewer than 50 occurrences each, creating a severe class imbalance characteristic of scientific multi-label classification.}
\label{fig:power-law}
\end{figure}

A critical characteristic of \textsc{AstroConcepts} is severe label imbalance, typical of real-world multi-label scenarios. Figure~\ref{fig:power-law} shows the label frequency distribution on a log-log scale, revealing a power-law pattern. The most frequent label (\textit{Galaxy evolution}) appears in 1,106 abstracts (5.1\%), while the median label appears in only 12 abstracts. We fit a power law $f(r) \propto r^{-\alpha}$ where $r$ is rank and $f(r)$ is frequency, obtaining $\alpha = 1.50$ with $R^2 = 0.825$.

The exponent $\alpha = 1.50$ indicates a moderately steep long-tail distribution. This suggests that \textsc{AstroConcepts} presents a substantial but not extreme class imbalance compared to other scientific classification scenarios. The moderate slope means mid-frequency (torso) labels retain more training examples than in steeper distributions; by rank 100, labels still average 58 examples, whereas in datasets with $\alpha = 2.0$, rank-100 labels would have only 25 examples. Nevertheless, a severe imbalance remains: 76\% of labels occur fewer than 50 times.

We partition labels into three frequency bins (see Table~\ref{tab:freq_bins}). Head labels (frequency $>$ 500): 17 labels (0.9\% of vocabulary) covering 12,288 assignments (13.1\% of total). These represent core concepts studied extensively across astrophysics subfields. Torso labels (50 $\leq$ frequency $\leq$ 500): 429 labels (23.0\%) covering 62,808 assignments (67.1\%). These represent moderately common research topics with sufficient training examples for standard supervised learning. Tail labels (frequency $<$ 50): 1,418 labels (76.1\%) covering 18,451 assignments (19.7\%). These represent specialized phenomena, emerging research areas, and niche methodologies with limited training examples, creating the zero-shot challenge we investigate in Section~\ref{experiments}.

\begin{table}[!h]
\centering
\scriptsize
\begin{tabular}{@{}lrrrr@{}}
\toprule
\textbf{Bin} & \textbf{\# Labels} & \textbf{\%} & \textbf{\# Assign.} & \textbf{\%} \\
\midrule
Head ($>$500) & 17 & 0.9 & 12,288 & 13.1 \\
Torso (50--500) & 429 & 23.0 & 62,808 & 67.1 \\
Tail ($<$50) & 1,418 & 76.1 & 18,451 & 19.7 \\
\midrule
\textbf{Total} & \textbf{1,864} & \textbf{100.0} & \textbf{93,547} & \textbf{100.0} \\
\bottomrule
\end{tabular}
\caption{Label frequency distribution across bins. The long tail contains 76\% of labels but only 20\% of assignments, creating a severe class imbalance characteristic of scientific multi-label classification.}
\label{tab:freq_bins}
\end{table}

This distribution creates the multi-level challenge characteristic of extreme multi-label classification: head labels are easily learned from abundant examples (hundreds per label), torso labels require careful modeling to generalize from moderate data (50--500 examples), and tail labels present a few-shot scenario (fewer than 50 examples, median: 12) where standard supervised methods struggle.

The most frequent labels span major astrophysics subfields: extragalactic astronomy (\textit{Galaxy evolution}, \textit{Active galactic nuclei}), stellar physics (\textit{Star formation}, \textit{Neutron stars}), planetary science (\textit{Exoplanets}, \textit{Exoplanet atmospheres}), and observational methods (\textit{Spectroscopy}, \textit{Astronomy data analysis}). No single subfield dominates the top-20 labels, confirming that \textsc{AstroConcepts} captures the breadth of modern astrophysics research rather than concentrating on narrow phenomena. This diversity is important for multi-label classification research: the corpus contains both well-represented core concepts and sparse specialized topics, enabling evaluation across the full spectrum of label frequencies.

\subsection{Concept Specificity Patterns}
Table~\ref{tab:depth_distribution} reveals that authors prefer moderately specific concepts (peaking at level 4). Both very general concepts (levels 1-2) and highly specialized terminology (levels 6+) are underrepresented relative to mid-level concepts. This concentration around moderate specificity creates additional challenges for classification systems beyond frequency imbalance. Models must handle vocabularies in which training examples cluster around mid-level concepts, with limited examples at both ends of the taxonomic spectrum, requiring systems to predict across varying specificity levels with highly imbalanced training signals.

\begin{table}[!h]
\centering
\begin{tabular}{@{}rrr@{}}
\toprule
\textbf{Depth} & \textbf{Count} & \textbf{\%} \\
\midrule
1 & 106 & 4.5 \\
2 & 106 & 4.5 \\
3 & 405 & 17.1 \\
4 & \textbf{680} & \textbf{28.7} \\
5 & 554 & 23.4 \\
6 & 341 & 14.4 \\
7 & 110 & 4.6 \\
8 & 36 & 1.5 \\
9 & 13 & 0.5 \\
10 & 11 & 0.5 \\
11 & 5 & 0.2 \\
\bottomrule
\end{tabular}
\caption{Distribution of UAT concept depths in \textsc{AstroConcepts} showing annotation preference patterns across taxonomic levels.}
\label{tab:depth_distribution}
\end{table}

\subsection{Concept Co-occurrence Patterns}
We analyzed how concepts co-occur in our corpus. We computed pointwise mutual information (PMI) for concept pairs, where PMI$(\ell_i, \ell_j) = \log \frac{P(\ell_i, \ell_j)}{P(\ell_i) P(\ell_j)}$ measures whether two concepts appear together more frequently than expected by chance. Positive PMI indicates stronger-than-expected co-occurrence, while negative PMI suggests mutual exclusivity. Focusing on pairs with positive PMI and sufficient co-occurrence ($\geq$ 10 abstracts), we found that authors rarely select hierarchically related concepts together. Concepts from unrelated subtrees show 2.3× higher average PMI than parent-child pairs, indicating that authors typically choose concepts spanning multiple taxonomic branches rather than selecting both general and specific terms from the same hierarchy. This cross-branch annotation behavior represents a challenge for classification systems, which must learn to predict conceptually diverse label combinations spanning the entire taxonomic structure.

\section{Experiments}
\label{experiments}
\subsection{Experimental Setup}

\paragraph{Task Formulation}
We formulate astrophysics concept classification as a multi-label task where, given an abstract $x$ concatenated with its title, the goal is to predict a subset $Y \subseteq \mathcal{L}$ of relevant UAT concepts from the complete label space $\mathcal{L}$ of 2,367 labels. We use title+abstract concatenation as input text to provide models with maximum available semantic information for concept prediction.

\paragraph{Data Partitioning}
We split the corpus into train (18,677 abstracts, 85\%) and test (3,025 abstracts, 15\%) sets using label-aware stratification. Labels appearing $\geq$15 times are stratified to achieve approximately 85/15 distribution per label, while labels with $<$15 occurrences are placed entirely in the training set to maximize training signal.

\paragraph{Evaluation Metrics}
Following standard practice in multi-label classification~\citep{zhang2014review,tsoumakas2010mining}, we evaluate using Macro-F1, which averages per-label F1 scores and is essential for imbalanced settings~\citep{wu2020revisiting}. For ranking evaluation, we use Precision/Recall at $k$ ($P@k$, $R@k$) with $k \in \{1, 3, 5\}$, chosen to align with the average number of assigned concepts per paper (4.31, see Table~\ref{tab:corpus_stats}).

\subsection{Baseline Approaches}
We establish baselines across multiple paradigms to understand effective modeling approaches for astrophysics concept classification, organized by increasing complexity: non-parametric methods, supervised neural models, and vocabulary-constrained LLMs.

\subsubsection{Non-Parametric Methods}
\label{non_parametric}
\paragraph{Rule-based Matching} We implement lexical matching based on the assumption that if a UAT concept is explicitly mentioned in the text, it should be assigned as a label. The method searches for exact string matches of UAT concept names within the title and abstract. For each concept, we check for its canonical designation and optionally include synonyms and abbreviations from the UAT taxonomy (e.g., searching for both \textit{active galactic nuclei} and \textit{AGN}). We evaluate two variants: Rule-based\textsubscript{w/o var} uses only canonical names, while Rule-based\textsubscript{w/ var} includes all alternative forms provided in the UAT.

\paragraph{$k$-Nearest Neighbors} This approach assumes abstracts with high contextual similarity should share similar concepts. We encode titles and abstracts using three embedding models: astroBERT~\citep{grezes-etal-astrobert} (adapted specifically for astrophysics texts), INDUS~\citep{Bhattacharjee-etal-indus} (a scientific language model covering astrophysics, earth science, and general physics), and \texttt{Qwen3-Embedding-8B}\footnote{\url{https://huggingface.co/Qwen/Qwen3-Embedding-8B}} (chosen for strong sentence similarity performance). For each abstract from the test set, we retrieve $k$ nearest training neighbors using cosine similarity, then predict the most frequent labels among them. We perform a grid search over $k \in \{5, 10, 20, 50\}$ and all embedding models, with detailed results provided in the appendix.
Table~\ref{tab:overall_performance} reports only the best-performing configuration (embedding model and $k$ value) for readability.

\subsubsection{Supervised Neural Models}
To investigate the effects of domain adaptation, we fine-tune three transformer models representing different levels of domain specialization: BERT~\citep{devlin-etal-2019-bert} (general-purpose), SciBERT~\citep{beltagy-etal-2019-scibert} (scientific domains), and astroBERT~\citep{grezes-etal-astrobert}. We add classification heads on $[CLS]$ representations and fine-tune end-to-end with max\_length=512, batch\_size=8, and AdamW optimizer. Due to computational resource constraints, we were unable to include Qwen models in the fine-tuning experiments. We perform a grid search over learning rates $\{2e-5, 3e-5, 5e-5\}$ and epochs $\{1, 2, 3, 5, 8, 10\}$ to find optimal hyperparameter configurations. Detailed results for each configuration are provided in the appendix. Table~\ref{tab:overall_performance} reports only the best-performing configuration (lr=2e-5, epochs=8) for readability.

\subsubsection{Vocabulary-Constrained LLMs}
\label{vocabulary-Constrained LLMs}
LLMs demonstrate strong capabilities on multi-label text classification tasks~\citep{zhou-etal-2024-quest,tabatabaei-etal-2025-large} but face challenges with large label spaces. Prompting an LLM directly to select from all 2,367 UAT concepts is infeasible due to context length limitations and label complexity. Preliminary experiments on a small subset yielded very poor results when using the complete label list, as the model hallucinated concepts or became overwhelmed by the extensive vocabulary. We therefore implement a two-stage approach: (1) use our best supervised model (astroBERT; see Table~\ref{tab:overall_performance}) to generate top-50 candidate labels for each abstract, (2) prompt \texttt{DeepSeek-V3-reasoner}~\citep{deepseek-v3} via its API\footnote{\url{https://api-docs.deepseek.com/}} to select relevant concepts from these candidates. We chose the top-50 threshold based on analysis showing that astroBERT's top-50 predictions covered approximately 82\% of ground-truth labels (see Figure~\ref{fig:coverage_analysis} in the appendix~\ref{app:label_coverage}), providing good coverage while maintaining manageable prompt length. We evaluate \texttt{DeepSeek-V3-reasoner} using the prompt shown in Figure~\ref{fig:llm_prompt} of the appendix~\ref{app:prompt}. This approach constrains the model to valid UAT terminology while leveraging its semantic understanding to select the most relevant concepts from the candidate set.

\section{Results and Analysis}
\label{results_analysis}
This section presents and analyzes results across all methods, revealing key insights about domain adaptation, frequency effects, and the comparative strengths of different learning paradigms for extreme multi-label scientific classification.

\subsection{Overall Performance and Domain Impact}

Table~\ref{tab:overall_performance} presents comprehensive results across all methods, revealing fundamental insights about learning paradigms for extreme multi-label scientific classification. 

\begin{table*}[!h]
\centering
\small
\begin{tabular}{lccccccccc}
\toprule
{}& \multicolumn{3}{c}{\textbf{Overall}} & \multicolumn{3}{c}{\textbf{Ranking Precision}} & \multicolumn{3}{c}{\textbf{Ranking Recall}} \\
\textbf{Method} & \textbf{Precision} & \textbf{Recall} & \textbf{F\textsubscript{1}} & \textbf{P@1} & \textbf{P@3} & \textbf{P@5} & \textbf{R@1} & \textbf{R@3} & \textbf{R@5} \\
\midrule
\multicolumn{8}{l}{\textit{\scriptsize{Non-parametric}}} \\
Rule-based\textsubscript{w/o var} & 0.2290 & 0.2130 & 0.1950 & $^\ddagger$ & $^\ddagger$ & $^\ddagger$ & $^\ddagger$ & $^\ddagger$ & $^\ddagger$ \\
Rule-based\textsubscript{w var} & 0.2170 & 0.2730 & 0.2140 & $^\ddagger$ & $^\ddagger$ & $^\ddagger$ & $^\ddagger$ & $^\ddagger$ & $^\ddagger$ \\
$k$-NN & 0.1165 & \textbf{0.7509} & 0.2017 & 0.6125 &0.4607& 0.3698 & 0.1398 & 0.3155 & 0.4221 \\
\midrule
\multicolumn{8}{l}{\textit{\scriptsize{Supervised neural models}}} \\
BERT & 0.1784 & 0.2273 & 0.1880 & 0.2975 & 0.2187 & 0.1784 & 0.0810 & 0.1730 & 0.2273 \\
SciBERT & 0.2020 & 0.2563 & 0.2127 & 0.3213 & 0.2409 & 0.2020 & 0.0872 & 0.1864 & 0.2563 \\
astroBERT&\textbf{0.3068}&0.3905 & 0.3243 & 0.4909 & 0.3733 & 0.3068 & 0.1360 & 0.2933 & 0.3905 \\
\midrule
\multicolumn{8}{l}{\textit{\scriptsize{Zero-shot prompting}}} \\
Deepseek&0.2891&0.6322& \textbf{0.3770} & \textbf{0.6502} & \textbf{0.4930} & \textbf{0.4017} & \textbf{0.1815} & \textbf{0.3837} & \textbf{0.5050} \\
\bottomrule
\end{tabular}
\caption{Overall performance on AstroConcepts test set. Best results shown in \textbf{bold}. $^\ddagger$Not applicable for rule-based methods.}
\label{tab:overall_performance}
\end{table*}

Our evaluation reveals a progression of insights across methodological paradigms. Simple rule-based matching achieves reasonable precision (0.229) but faces fundamental limitations: string matching of concept mentions may not reflect the core research focus. Common terms like "\textit{photon}" appear across diverse papers, while implicit language and incomplete variant coverage constrain effectiveness.

Moving to similarity-based approaches, $k$-NN achieves exceptional recall (0.751) but poor precision (0.117). Crucially, domain-adapted astroBERT embeddings outperform general models like Qwen, better capturing subtle concept distinctions essential for astrophysics. However, performance degrades beyond $k=10$ neighbors as noise from irrelevant papers accumulates.

The most striking finding emerges with vocabulary-constrained DeepSeek, which outperforms even the best domain-adapted model (astroBERT) by 16\% in F\textsubscript{1} score (0.377 vs 0.324). This demonstrates that domain expertise can be effectively incorporated through vocabulary constraints rather than solely through model parameters. Meanwhile, supervised domain adaptation shows clear value: astroBERT substantially outperforms SciBERT (0.324 vs 0.213) with improvements concentrated in precision and confident prediction rather than comprehensive recall.

Together, these results reveal that effective scientific text classification benefits from hybrid approaches combining general language understanding with structured domain knowledge, opening promising directions for parameter-efficient scientific NLP.

\subsection{The Long-Tail Challenge}
The extreme label distribution in \textsc{AstroConcepts} (78\% of concepts have $<50$ examples) enables the systematic analysis of long-tail performance patterns. Table~\ref{tab:frequency_analysis} presents frequency-stratified results across Head ($>500$ examples), Torso ($50$--$500$), and Tail ($<50$) concepts, revealing insights that aggregate metrics cannot capture.

\begin{table*}[!h]
\centering
\small
\begin{tabular}{r|ccccccccc|c}
\toprule
& \multicolumn{3}{c}{\textbf{F\textsubscript{1}}} & \multicolumn{2}{c}{\textbf{Head}} & \multicolumn{2}{c}{\textbf{Torso}} & \multicolumn{2}{c|}{\textbf{Tail}} & \\
\textbf{Method} & \textbf{Head} & \textbf{Torso} & \textbf{Tail} & \textbf{P@3} & \textbf{R@3} & \textbf{P@3} & \textbf{R@3} & \textbf{P@3} & \textbf{R@3} & $\Delta^*$ \\
\midrule
BERT & 0.180 & 0.167 & 0.021 & 0.084 & 0.205 & 0.168 & 0.183 & 0.012 & 0.024 & 0.159 \\
SciBERT & 0.193 & 0.197 & 0.023 & 0.084 & 0.211 & 0.199 & 0.212 & 0.015 & 0.026 & 0.170 \\
astroBERT & 0.216 & 0.312 & 0.081 & 0.092 & 0.230 & 0.311 & 0.333 & 0.046 & 0.088 & 0.135 \\
DeepSeek & \textbf{0.243} & \textbf{0.376} & \textbf{0.198} & \textbf{0.107} & \textbf{0.266} & \textbf{0.417} & \textbf{0.443} & \textbf{0.135} & \textbf{0.267} & \textbf{0.045} \\
\bottomrule
\end{tabular}
\caption{performance across frequency bins. Head/Torso/Tail thresholds: $>500$/$50$--$500$/$<50$ training examples. $^*$$\Delta$ = Head F\textsubscript{1} - Tail F\textsubscript{1} (lower is better). Best results in \textbf{bold}.}
\label{tab:frequency_analysis}
\end{table*}

Three patterns emerge, providing useful insights into handling extreme imbalance. First, domain adaptation provides asymmetric benefits: astroBERT shows modest Head improvements over SciBERT (0.193 $\rightarrow$ 0.216) but larger relative gains for Tail concepts (0.023 $\rightarrow$ 0.081), though absolute performance remains limited across traditional approaches. Second, all methods exhibit a consistent "torso peak" where mid-frequency concepts achieve optimal performance. This pattern suggests fundamental properties of scientific vocabulary learning, in which models balance sufficient training signal with complexity issues at frequency extremes. Most significantly, the vocabulary-constrained approach achieves superior tail performance (F\textsubscript{1}: 0.198 vs 0.081 for astroBERT), demonstrating that domain expertise can be effectively incorporated through structured constraints rather than parameter fine-tuning alone. This hybrid approach, combining general language understanding with domain-specific vocabulary guidance, proves particularly effective for rare concepts. Finally, we introduce frequency robustness ($\Delta$ = Head F\textsubscript{1} - Tail F\textsubscript{1}) as a critical evaluation dimension. The constrained approach achieves 3× better robustness than SciBERT ($\Delta=0.045$ vs 0.170), indicating that architectural choices and constraint mechanisms matter more for handling frequency imbalance than domain specialization alone.

\section{Discussion}
\label{discussion}
Our systematic evaluation reveals fundamental insights into extreme multi-label classification in scientific domains while establishing new evaluation paradigms that advance understanding beyond existing benchmarks.

\paragraph{Addressing the Research Questions}
Our findings provide clear answers to the three research questions posed. Regarding RQ1 (handling extreme imbalance), vocabulary-constrained LLMs demonstrate superior robustness across frequency bins, achieving 3× better head-tail balance than traditional supervised approaches. For RQ2 (domain adaptation benefits), we find asymmetric improvements that focus on rare, specialized terminology rather than on frequent concepts, challenging assumptions about uniform domain adaptation effects. RQ3 (competitive LLM performance) is answered affirmatively: the hybrid approach combining astroBERT candidate generation with LLM selection achieves competitive results while requiring substantially fewer computational resources than full fine-tuning.

\paragraph{Methodological Impact}
We establish frequency-stratified evaluation as essential for extreme multi-label assessment and introduce the head-tail gap ($\Delta$) as a robustness metric that reveals performance patterns invisible in aggregate scores. The systematic comparison across paradigms demonstrates that different approaches excel in complementary areas: rule-based methods provide interpretability but limited coverage, k-NN offers high recall with domain-adapted embeddings, supervised models achieve confident predictions, and constrained LLMs balance precision-recall trade-offs effectively.

\paragraph{Broader Implications}
The torso peak phenomenon, in which all methods perform best on mid-frequency concepts, suggests fundamental limits to learning from extreme imbalance that transcend architectural choices. This finding has immediate implications for resource allocation in scientific NLP: optimization efforts should target the frequency regions where improvement is most feasible, rather than aiming for uniform performance gains across all concepts.

\paragraph{Relationship to Complementary Resources}
A related effort from~\citet{ting-etal-2025-astromlab} extracted 9,999 concepts from 408,590 astrophysics papers using LLM-based pipelines and clustering over full-text content. The two resources address different but complementary needs. \textsc{AstroMLab~5} produces a semantically rich, emergent vocabulary optimized for discovery, knowledge graph construction, and temporal analysis at scale. \textsc{AstroConcepts}, by contrast, provides controlled-vocabulary annotations grounded in the UAT, enabling reproducible evaluation of classification methods under extreme label imbalance, a use case for which a fixed label space, train/test split, and expert-assigned labels are essential. Looking forward, the two resources open natural avenues for joint investigation: \textsc{AstroMLab~5} concepts could serve as additional candidate labels or weak supervision signals for tail concepts in \textsc{AstroConcepts}, while UAT-grounded annotations could provide an extrinsic evaluation signal for the quality of LLM-extracted concept vocabularies.

\paragraph{Limitations and Future Directions}
Our findings are domain-specific and require validation across other scientific fields before broader claims about scientific NLP can be established. The persistent tail performance challenge (best F\textsubscript{1}: 0.198) indicates fundamental limitations in current approaches, suggesting opportunities for architectural innovations tailored to extreme imbalance scenarios. Future work should explore integrating structured domain knowledge with few-shot learning approaches. An important validation step is to audit a stratified sample of the corpus through expert review and inter-annotator agreement analysis, which would quantify annotation noise and its downstream effect on recall-based metrics, particularly for tail concepts where incomplete author-assigned labels may inflate the apparent difficulty. Extending this methodology to other scientific domains is essential to establish whether the torso-peak phenomenon, asymmetric domain adaptation benefits, and the effectiveness of vocabulary-constrained approaches generalize beyond astrophysics. Finally, more extensive experiments with the LLM filtering stage are needed: evaluating DeepSeek over candidate sets generated by BERT and SciBERT, in addition to astroBERT, would isolate the contribution of the candidate generator's quality from the LLM's own re-ranking ability, providing a cleaner assessment of where the performance gains originate.

\section{Conclusion}
\label{conclusion}

\textsc{AstroConcepts} provides the NLP community with essential resources for investigating extreme multi-label classification in scientific domains, in particular for astrophysics. Through systematic evaluation across traditional, neural, and vocabulary-constrained approaches, we demonstrate three key insights that advance understanding of scientific text classification. The effectiveness of hybrid vocabulary-constrained approaches, in which astroBERT generates candidate labels and DeepSeek selects from this constrained set, demonstrates that domain expertise can be incorporated through structured vocabulary guidance rather than through extensive LLM fine-tuning. This approach achieves competitive performance (F\textsubscript{1}: 0.377) while requiring only inference costs for the domain model and API calls for the LLM, opening promising directions for cost-effective scientific NLP that combines specialized knowledge extraction with general language understanding capabilities. Domain adaptation benefits concentrate asymmetrically on rare, specialized terminology, suggesting that specialized models primarily handle concepts beyond general model capabilities rather than improving performance uniformly. Our frequency-stratified evaluation framework, combined with robustness metrics, provides useful methods for assessing extreme multi-label systems in which aggregate scores can mask critical performance patterns. These contributions address a critical methodological gap by enabling systematic evaluation of extreme imbalance while providing actionable insights for scientific NLP practitioners. The corpus, baselines, and evaluation framework lay the foundations for future research on specialized domain classification, while our findings on vocabulary-constrained approaches indicate promising directions for resource-efficient scientific text processing systems. To facilitate future research, we make the \textsc{AstroConcepts} corpus publicly available. The persistent challenges in tail performance underscore opportunities for novel approaches that integrate structured knowledge with text-based classification. Future work should prioritize annotation validation through expert audits and inter-annotator agreement studies, systematic cross-domain replication, and controlled ablations of the LLM re-ranking stage across candidate generators of varying quality. As the scientific literature continues to expand and specialized terminology increases, effective handling of extreme imbalance becomes essential for scientific NLP applications.

\section{Ethical Considerations}
All abstracts are from published scientific papers that are publicly accessible. We include only bibliographic metadata (bibcode, title, abstract, publication year, journal) and assigned UAT concepts.

\section{Bibliographical References}\label{sec:reference}

\bibliographystyle{lrec2026-natbib}
\bibliography{bibliography}


\appendix

\section*{A.1. Complete Prompt Template (Section~\ref{vocabulary-Constrained LLMs})}
\label{app:prompt}
Figure~\ref{fig:llm_prompt} shows the designed prompt as part of our experiments.

\begin{figure*}[!h]
\centering
\fcolorbox{gray!30}{gray!5}{\begin{minipage}{0.9\textwidth}
\small
\textcolor{blue!70!black}{\texttt{You are an expert astrophysicist and scientific topic classifier.}}\\
\textcolor{blue!70!black}{\texttt{Your task is to choose between 1 to 10 labels from the candidate list that}}\\
\textcolor{blue!70!black}{\texttt{accurately describe the main scientific themes of the following research paper.}}\\
\textcolor{blue!70!black}{\texttt{A label should be selected only if it is clearly relevant to the paper's content.}}\\
\\
\textcolor{gray!60}{\texttt{---}}\\
\textcolor{teal!70!black}{\textbf{\texttt{Abstract:}}}\\
\textcolor{gray!70}{\texttt{\{abstract\}}}\\
\\
\textcolor{gray!60}{\texttt{---}}\\
\textcolor{teal!70!black}{\textbf{\texttt{Candidate topics suggested by the model:}}}\\
\textcolor{gray!70}{\texttt{\{topk\_labels\}}}\\
\\
\textcolor{purple!70!black}{\texttt{Return your answer in valid JSON as follows:}}\\
\textcolor{orange!70!black}{\texttt{\{}}\\
\textcolor{orange!70!black}{\texttt{~~"selected\_labels": ["label1", "label2", ...]}}\\
\textcolor{orange!70!black}{\texttt{\}}}\\
\textcolor{red!70!black}{\texttt{Do not include explanations or text outside the JSON.}}
\end{minipage}}
\caption{Prompt template used for vocabulary-constrained LLM classification.}
\label{fig:llm_prompt}
\end{figure*}

\section*{A.2. astroBERT label coverage (Section~\ref{vocabulary-Constrained LLMs})}
\label{app:label_coverage}
\begin{figure*}[!h]
    \centering
    \includegraphics[scale = 0.45]{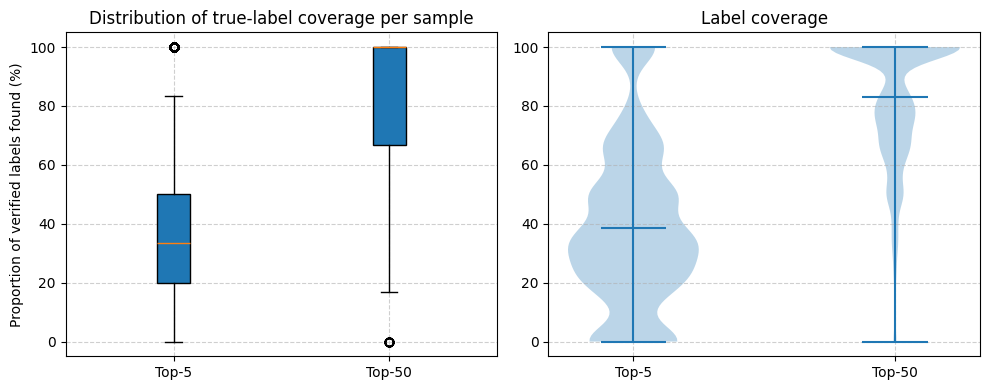}
    \caption{Fine-tuned astroBERT Label Coverage}
    \label{fig:coverage_analysis}
\end{figure*}

Figure~\ref{fig:coverage_analysis} shows that astroBERT's top-50 predicted labels covers approximately 82\% of the ground-truth labels.

\end{document}